\documentclass[11pt,a4paper]{article}
\usepackage[hyperref]{emnlp2020}
\usepackage{times}
\usepackage{latexsym}

\usepackage{microtype}

\usepackage{amsmath, amscd, amssymb, amsthm}
\usepackage{graphicx}
\usepackage{bm}
\usepackage{bbm}
\usepackage{booktabs}
\usepackage{enumitem}
\usepackage{footnote}
\usepackage{mathtools}
\usepackage{multicol}
\usepackage{multirow}
\usepackage{natbib}
\usepackage{outlines}
\usepackage{pifont}% http://ctan.org/pkg/pifont
\usepackage{subcaption}
\usepackage{soul}
\usepackage{tabularx}
\usepackage{tabulary}
\usepackage{url}
\usepackage{xcolor}
\usepackage[utf8]{inputenc}
\usepackage{hyperref}
\usepackage{cleveref}
\usepackage{verbatim}

\aclfinalcopy % Uncomment this line for the final submission
\title{Learning Visual-Semantic Embeddings for\\ Reporting Abnormal Findings on Chest X-rays}

\author{
  Jianmo Ni\thanks{\quad Now at Google}, Chun-Nan Hsu, Amilcare Gentili, Julian McAuley \\
  University of California, San Diego \\
  \texttt{ \{jin018,chunnan,agentili,jmcauley\}@ucsd.edu } \\} 

\date{}

\begin{document}
\maketitle
\begin{abstract}
% Automatic medical image report generation has drawn growing attention due to its potential to alleviate radiologists' workload.
% Existing work on report generation often trains encoder-decoder networks to generate complete reports. 
% % We discover that such models are prone to be affected by various data bias (e.g.~label imbalance,
% However, such models are affected by data bias (e.g.~label imbalance) and face common issues inherent in text generation models (e.g.~repetition). 
% In this work, we focus on reporting abnormal findings on
% radiology images;
% % chest X-ray (CXR) images. 
% % We identify abnormal findings from radiology reports and consider them as a training set instead of the complete reports.
% %JIANMO: Emphasize we don't use the complete data but only the abnormal ones -- our main contribution.
% instead of training on complete radiology reports, we propose a method to identify abnormal findings from the reports in addition to grouping them with unsupervised clustering and minimal rules.
% We propose Conditional Visual-Semantic Embeddings
% to align images and fine-grained abnormal findings in a joint embedding space.
% We demonstrate that our method is able to retrieve abnormal findings and outperforms existing generation models on
% % clinical accuracy. 
% both clinical correctness and text generation metrics.

Automatic medical image report generation has drawn growing attention due to its potential to alleviate radiologists' workload. Existing work on report generation often trains encoder-decoder networks to generate complete reports. However, such models are affected by data bias (e.g.~label imbalance) and face common issues inherent in text generation models (e.g.~repetition). In this work, we focus on reporting abnormal findings on radiology images; instead of training on complete radiology reports, we propose a method to identify abnormal findings from the reports in addition to grouping them with unsupervised clustering and minimal rules. We formulate the task as cross-modal retrieval and propose Conditional Visual-Semantic Embeddings to align images and fine-grained abnormal findings in a joint embedding space. We demonstrate that our method is able to retrieve abnormal findings and outperforms existing generation models on both clinical correctness and text generation metrics.

\end{abstract}

\section{Introduction}
\label{chapter6:sec:intro}

Understanding abnormal findings on radiographs (e.g.~chest X-Rays) is a crucial task for radiologists. %while the process is often time-consuming and prone-to-error. 
% but is a time consuming and error-prone process.
%Researchers have studied on 
There has been growing interest in
automatic radiology report generation to alleviate the workload of radiologists and improve patient care. Following the success of neural network models in image-to-text generation tasks (e.g.~image captioning), researchers have trained CNN-RNN encoder-decoder networks to generate reports given radiology images \citep{Shin2016LearningTR,Kougia2019ASO}. 

\begin{figure*}[t]
    \centering
    \includegraphics[width=\linewidth]{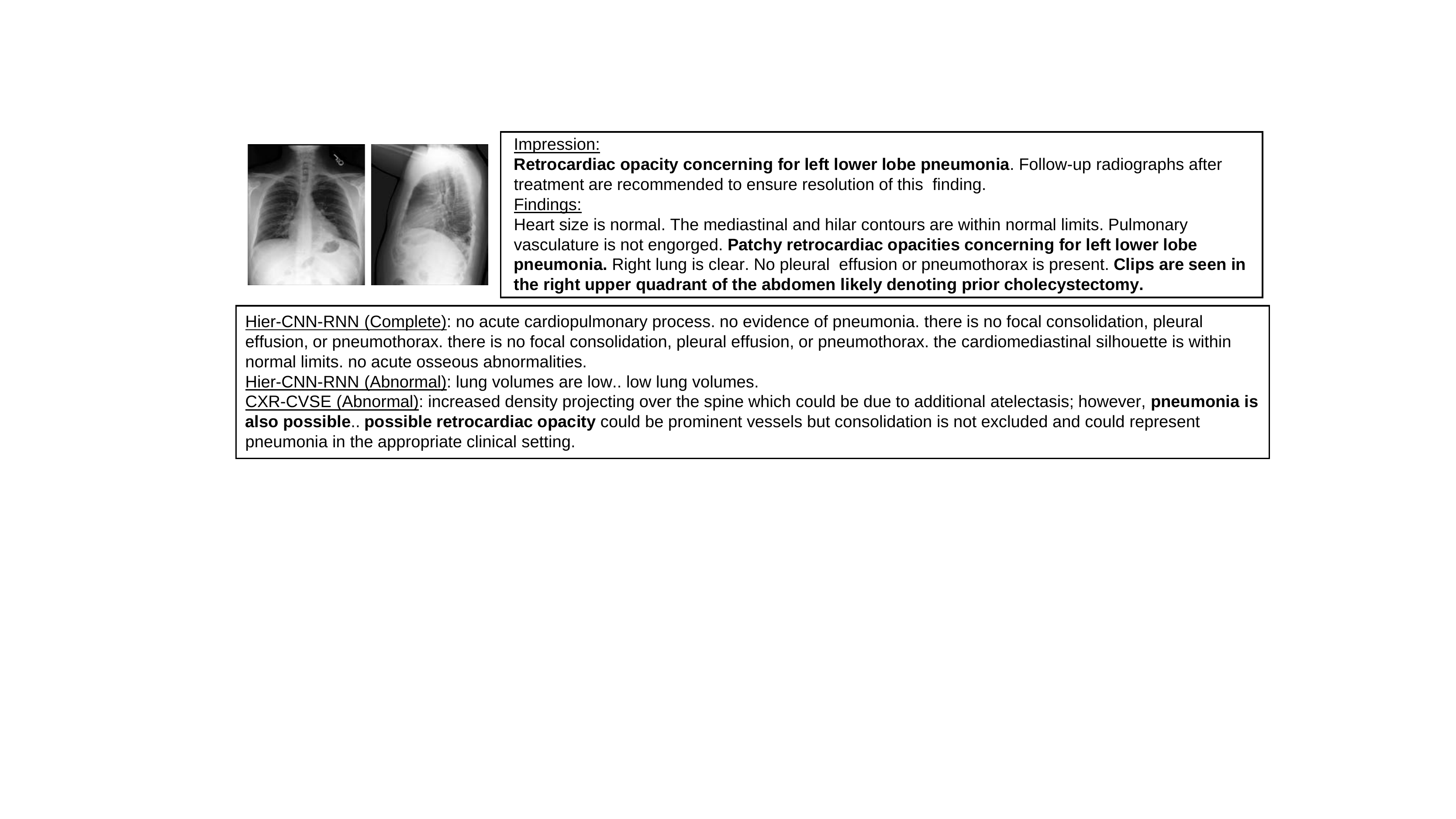}
    \caption{Example of CXR images (frontal and lateral views) and the associated report. Bolded are abnormal findings in the ground-truth and predictions. The CNN-RNN model trained on the complete reports tends to generate normal findings. Both CNN-RNN models generate repetitive sentences. }
    \label{fig:bias}
    % \vspace{-0.5\baselineskip}
\end{figure*}

Although such models are able to generate fluent reports, the generation quality is often limited by biases introduced from training data or the training process. \Cref{fig:bias} shows an example of chest X-rays (CXRs) and the associated reports from a public dataset \citep{Johnson2019MIMICCXRAL}, along with the outputs generated by different models.%
\footnote{For a CXR report, `Findings' is a detailed description and the `Impression' is a summary.}  
One issue is that models trained on complete reports tend to generate normal findings as they dominate the dataset \citep{AddressingDataBias};
% However, identifying abnormal findings (which are rare in the data) might be of higher importance to radiologists. 
%On the other hand, 
another issue is that
such generation models 
%have difficulties 
struggle
to generate long and diverse reports as in other natural language generation (NLG) tasks \citep{Boag2019BaselinesFC}.

% Typical sources of bias include:
% %of biases include:
% (1) \emph{data bias}, such as the imbalance between normal findings and abnormal findings, and
% (2) \emph{exposure bias} during the training of the decoder using the teacher-forcing strategy.
% The former will encourage the model to generate normal findings, as the model sees them more frequently on the training set \citep{AddressingDataBias}. 
% % The latter will make the model perform poorly at the inference time once the generation diverges from the training data. As a result, these models tend to generate repetitive results
% The latter is a common issue in natural language generation tasks, which leads to repetitive outputs \citep{Liu2019ClinicallyAC}. \Cref{fig:bias} shows example outputs of the CNN-RNN models, which demonstrates the affect of such biases.

In this work, we focus on reporting abnormal findings on radiology images which are of higher importance to radiologists. To address issues of data bias, we propose a method to identify abnormal findings from existing reports and further use K-Means plus minimal mutual exclusivity rules to group these abnormal findings, which reduces the substantial burden of curating templates of abnormal findings.
% image-to-text retrieval 
% Instead of training report generation models, we seek to approach the task using a retrieval-based method that is able to capture relevant abnormal findings from the given CXR images.
%JIANMO: Explain why we formulate it as a cross-modal retrieval task. Probably need to rephrase.
Given the fact that radiology reports are highly similar and have a limited vocabulary \citep{Gabriel2018IdentifyingAC}, 
we propose a cross-modal retrieval method to capture relevant abnormal findings from radiology images.
Our contributions are summarized as:
\setlist{nolistsep}
\begin{itemize}[noitemsep]
    \item We learn conditional visual-semantic embeddings on radiology images and reports, which can be used to measure the similarity between image regions and abnormal findings by optimizing a triplet ranking loss.
    \item We develop an automatic approach to identify and group abnormal findings from large collections of radiology reports.
    \item We conduct comprehensive experiments to show that our retrieval-based method trained on the abnormal findings largely outperforms encoder-decoder generation models on clinical correctness and NLG metrics.
\end{itemize}

\section{Related Work}
\label{chapter6:sec:related_work}
\subsection{Hierarchical encoder-decoder models}
\citet{Jing2017OnTA} proposed a co-attention based Hierarchical CNN-RNN model that jointly trains two tasks: report generation and Medical Text Indexer (MTI) prediction. The model first predicts MTI tags and the semantic embeddings of the predictions are fed into the cascaded decoder for generation.
Similarly, \citet{Yuan2019AutomaticRR} extracted medical concepts from the CXR reports using SemRep\footnote{\url{https://semrep.nlm.nih.gov/}} as alternatives to MTI tags.
% They further explore several type of fusions to incorporate the semantic medical concepts' semantic embeddings into the decoder.
To address data bias, \citet{AddressingDataBias}
proposed a CNN-RNN model with dual word-level decoders: one for abnormal findings and the other for normal findings. It jointly predicts whether the next sentence is a normal or abnormal finding, and uses the corresponding decoder to generate the next sentence. 
However, it still formulates the task as text generation and has the limitations of such models. 

\subsection{Hybrid retrieval-generation models}
There has been increasing interest in studying hybrid retrieval-generation models to complement generation.
% The retrieval module is expected to find relevant template sentences that dominating the data and the generation model is trained to handle the rare cases with high variance in the data.
\citet{Li2018HybridRR} introduced a hybrid retrieval-generation framework which decides at each step whether it retrieves a template or generates a sentence. % The policy is trained via reinforcement learning.
\citet{Li2019KnowledgedrivenER}
proposed a model based on abnormality graphs, which first predicts existing abnormalities on the radiology images, then retrieves and paraphrases the templates of that abnormality. 
However, such models usually require non-trivial human effort to construct high quality prior knowledge (e.g.~sentence templates, abnormality terms). 
Unlike previous work, we leverage unsupervised methods and minimal rules to group sentences into different abnormality clusters, seeking to minimize human effort.

% \subsection{Clinical accuracy-aware models}
% \Citet{Liu2019ClinicallyAC} show that existing encoder-decoder networks focus on optimizing the natural language metrics (e.g. BLEU, CIDEr) of the generated report does not guarantee clinical accuracy. To mitigate this issue, it develops a CNN-RNN model that takes the clinical accuracy as reward and applies reinforcement learning to directly optimize it.  

\subsection{Visual-semantic embeddings for cross-modal retrieval}
Learning visually grounded semantics to facilitate cross-modal retrieval (i.e.,~image-to-text and text-to-image) is a challenging task for cross-modal learning \citep{Faghri2018VSEIV,Wu2019UnifiedVE}. 
Different from image captioning tasks, radiology reports are often longer and consist of multiple sentences, each related to different abnormal findings; meanwhile, there are fewer
distinct objects in radiology images and the differences among images are more subtle. 
% To this end, we propose a conditional visual-semantic embeddings that learn an embedding for each abnormal finding rather than a holistic embedding for the complete report.

\section{Approach}
\label{chapter6:sec:approach}

% \subsection{Background}
Given radiology images $I_f$ and $I_l$ from the frontal and lateral view, 
%JIANMO: here's the definition for the conventional model
Hierarchical CNN-RNN based methods predict complete medical reports $R=\{s_1,s_2,\ldots,s_N\}$, consisting of $N$ sentences. Each sentence $s_i$ is generated hierarchically:
\begin{equation}
\label{eq:cnn-rnn}
    P(s_i) = \prod_{t=1}^{T_i} P(w_i^t | w_i^{<t}, s_{<i}, E_f, E_l),
\end{equation}
where $E_f$ and $E_l$ are the feature maps of the images $I_f$ and $I_l$ generated by the CNN encoder, and $w_i^t$ is the $t$-th word at the $i$-th sentence. 

Instead of training such generation models, we approach the task as a cross-modal retrieval method.
In particular, we propose a model that (1) measures the similarity between images and abnormal findings, and (2) identifies fine-grained relevant image regions for each abnormal finding.

\subsection{Problem definition}
\label{subsec:definition}
Assume each report $R_a=\{a_1,a_2,\ldots,a_M\}$ includes $M$ abnormal findings (i.e.,~sentences). $R_a$ is a subset of the complete report $R=\{s_1,s_2,\ldots,s_N\}$, where $s_i$ can either be an abnormal sentence $a_i$ or not. 

Let $v \in \mathbb{R}^{d1}$ be the semantic embedding of an abnormal finding $a$ of this report, and $E=\{m_j \in \mathbb{R}^{d2}\}_{j=1}^{w \times h}$ be the feature maps of the radiology image $I$ associated with $R_a$, where $j$ means the $j$-th region of the feature map. We first transform them into the joint embedding space $\mathbb{R}^{d}$ with separate linear projection layers:
\begin{equation*}
    \mathbf{v} = \mathrm{norm}(\mathrm{linear}(v)); \mathbf{m}_j = \mathrm{norm}(\mathrm{linear}(m_j)),
\end{equation*}
% $\mathbf{v} = \mathrm{norm}(\mathrm{linear}(v))$ and $\mathbf{m}_j = \mathrm{norm}(\mathrm{linear}(m_j))$,
where we apply $l_2$ normalization on the joint embeddings to improve training stability, following work in visual-semantic embeddings \citep{Faghri2018VSEIV}.

Next, we need to measure the similarity between the semantic and visual embeddings. As different regions may include details about different abnormal findings, we propose Conditional Visual-Semantic Embeddings (CVSE) to learn the fine-grained matching between regions and a target abnormal finding:
\begin{equation}
\label{eq:attention}
\begin{aligned}
    d(a, I) & = - \sum_{1 \le j \le w \times h}{\alpha_{j} ||\mathbf{m}_j - \mathbf{v}||^2}, \\
    \hat{\alpha_{j}} & = {\mathbf{v}_{\alpha}}^{\top} (W_{\alpha}[ \mathbf{m}_j ; \mathbf{v} ]+\mathbf{b}_{\alpha}), \\ 
    \bm{\alpha} & = \mathrm{softmax}(\bm{\hat{\alpha}}),
\end{aligned}
\end{equation}
where $\alpha_{j}$ is the attention score that represents the relevance between the region $\mathbf{m}_j$ and the abnormal finding $\mathbf{v}$, $d(a, I)$ is the similarity score between image $I$ and the abnormal finding $a$, which is calculated as an attention-weighted sum over the similarity scores of each region with the abnormal finding. We use the 
%minus 
(negative)
squared $l_2$ distance to measure similarity. Since each report has both frontal and lateral views, the final similarity score is calculated as the average: 
\begin{equation}
    d_{*}(a, I) = \frac{1}{2}(d(a, I_f) + d(a, I_l)).
\end{equation}

Finally, we optimize the hinge-based triplet ranking loss to learn the visual-semantic embeddings:
\begin{equation}
\begin{aligned}
    % \mathcal{L} & = \sum_{I} [ d_{*}(a^{+}, I) - d_{*}(a^{-}, I) + \delta]_{+} \\
    % & + \sum_{a} [ d_{*}(a, I^{+}) - d_{*}(a, I^{-}) + \delta]_{+},
    \mathcal{L} & = \sum_{I} [ d_{*}(a^{-}, I) - d_{*}(a^{+}, I) + \delta]_{+} \\
    & + \sum_{a} [ d_{*}(a, I^{-}) - d_{*}(a, I^{+}) + \delta]_{+},
\end{aligned}
\end{equation}
where $\delta$ is the margin, $[x]_{+} = \mathit{max}(x, 0)$ is the hinge loss, $a^{+}$ ($I^{+}$) denotes a matched abnormal finding (image) from the training set while $a^{-}$($I^{-}$) denotes an unmatched abnormal finding (image) sampled during training. 

\subsection{Extracting and clustering abnormal findings}

To identify abnormal findings in radiology reports, we train a sentence-level classifier which determines whether a sentence includes abnormal findings or not. We fine-tuned BERT \citep{Devlin2019BERTPO} on an annotated sentence-level dataset released by \citet{AddressingDataBias}, which is a labeled subset of the Open-I dataset \citep{openi}. 
We achieve an F1-score of 98.3 on the held-out test set. We then use it to distantly label the reports from the MIMIC-CXR dataset \citep{Johnson2019MIMICCXRAL}, which is the largest public CXR imaging report dataset.
% \footnote{We sampled sentences from the labeled MIMIC-CXR dataset and manually check the quality.}

%TODO: need to emphasize many sentences are derived from templates, thus we can use such retrieval based method
Given that most medical reports are written following certain templates, many abnormal findings are often paraphrases of each other. We obtain the sentence embeddings via pre-trained models and apply K-Means to cluster the sentences about similar abnormal findings into 500 groups. We also design several simple mutual exclusivity rules to refine the groupings. We consider critical attributes such as position (e.g.~left, right), severity (e.g.~mild, severe) which often are not present at the same time. Then we apply these rules to separate each group formed by K-Means. Ultimately, we obtained 1,306 groups of abnormal findings.

\begin{table*}[t]
\centering
\small
\caption{Comparisons of different models' clinical accuracy and NLG metrics. Accuracy, precision and recall are the macro-average across all 14 diseases.}
\vspace{-0.5\baselineskip}
\label{tbl:clinical_accuracy}
\begin{tabularx}{\linewidth}{X r r r | r r r r}
\toprule
Model & Accuracy & Precision & Recall & BLEU-4 & BLEU-1 & ROUGE-L & METEOR \\ 
\midrule
MIMIC-CXR (Abnormal) &  &  &  &  &  &  &  \\
CVSE + mutual exclusivity	& \textbf{0.863}	& \textbf{0.317}	&  \textbf{0.224} & \textbf{0.036} & 0.192 & \textbf{0.153} & 0.077 \\
CVSE	& 0.856	& 0.303	&  0.218 & 0.032 & \textbf{0.197} & 0.153 & \textbf{0.088} \\
Hier-CNN-RNN & 0.850 & 0.261 & 0.157 & 0.019 & 0.084 & 0.149 & 0.059 \\
Hier-CNN-RNN + shuffle & 0.853	& 0.172	& 0.117 & 0.013 &	0.064 & 0.130 & 0.046 \\
\midrule
MIMIC-CXR (Complete) &  &  &  &  &  &  &  \\
Hier-CNN-RNN + complete & 0.835 & 0.145 & 0.135 & 0.096 & 0.258 & 0.257 & 0.121 \\
Hier-CNN-RNN + co-attention	& 0.843	& 0.156	& 0.127 & 0.098 & 0.281 & 0.252 & 0.120 \\
Hier-CNN-RNN + dual	& 0.843	& 0.194	& 0.142 & 0.095 & 0.282 & 0.256 & 0.123 \\ 
\bottomrule
\end{tabularx}
\end{table*}

\section{Experiments}
\label{chapter6:sec:experiments}
We compare CVSE with the state-of-the-art report generation models and simple baseline models to answer two research questions---\textbf{RQ1}: Does our retrieval-based method outperform generation models? \textbf{RQ2}: Do the visual-semantic embeddings capture abnormal findings grounded on images?

\subsection{Baselines}
We consider (1) the Hier-CNN-RNN model \citep{Jing2017OnTA,Liu2019ClinicallyAC}, as denoted in \cref{eq:cnn-rnn}; (2) Hier-CNN-RNN + co-attention \citep{Jing2017OnTA} with co-attention on both the images and the predicted medical concepts;
% \footnote{We use SemRep (i.e.~a UMLS-based program) to extract 93 highly frequent medical concepts from the training set.} 
(3) Hier-CNN-RNN + dual, with the dual word-level decoders \citep{AddressingDataBias}. We also implement two simple variants: (4) Hier-CNN-RNN + complete, which considers the complete medical reports (i.e.,~both normal and abnormal findings) as input;  (5) Hier-CNN-RNN + shuffle, whose input reports have a shuffled sentence order. \citet{Vinyals2015OrderMS} has shown that input order
affects the
performance for encoder-decoder models and 
%this baseline 
(5) could potentially address the training issue due to the static input order.

In all experiments, the abnormal set and complete set consist of the same (image, report) pairs. As discussed in \Cref{subsec:definition}, the abnormal set only considers the abnormal finding sentences of the report, which is a subset of sentences of the complete report. We compare these two sets to show that models trained on the abnormal sentences would achieve substantial improvement than those trained on the complete reports, which has not been studied before.

We use the CheXpert labeler to evaluate the clinical accuracy of the abnormal findings reported by each model, which is the state-of-the-art medical report labeling system \citep{CheXpertAL,Johnson2019MIMICCXRAL}. 
Given sentences of abnormal findings, CheXpert will give a positive and negative label for 14 diseases. We then calculate the Precision, Recall and Accuracy for each disease based on the labels obtained from each model's output and from the ground-truth reports.

\subsection{Implementation details}
We consider CXRs from the MIMIC-CXR dataset with both frontal and lateral views 
which include at least one abnormal finding. 
Ultimately, we obtain 26,946/3,801/7,804 CXRs for the train/dev/test sets, respectively.
For the CVSE model, we set $\alpha$ to 0.2 and for each sample we randomly pick 8 negative samples. We use the pre-trained DenseNet-121 to obtain the feature maps of the CXR images. We use the pre-trained biomedical sentence embeddings \citep{Zhang2019BioWordVecIB} to obtain initial embeddings for the abnormal findings.\footnote{https://github.com/ncbi-nlp/BioSentVec} The final dimension of the joint embedding $d$ is set to 512. We take the top 3 retrieval results as the predicted abnormal findings. 
For all CNN-RNN based models, we use a VGG-19 model as the encoder, a 1-layer LSTM as the sentence decoder and a 2-layer LSTM as the word decoder.
% \footnote{We observe a slightly better performance of VGG-19 than DenseNet-121 for the generation models.} 
All dimensions are set to 512.
Greedy search is applied during the decoding stage, following \citet{Jing2017OnTA}. Our code are available online.\footnote{https://github.com/nijianmo/chest-xray-cvse}

\begin{figure*}[t]
    \centering
    \includegraphics[width=0.9\linewidth]{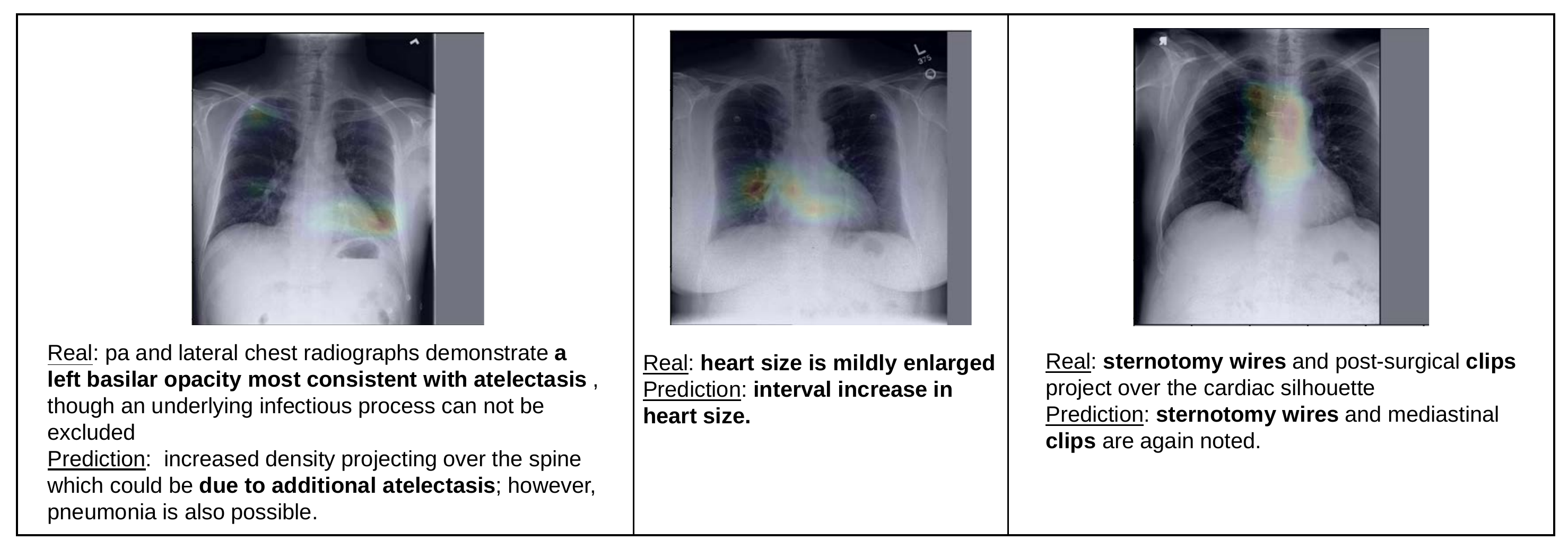}
    % \vspace{-0.5\baselineskip}
    \caption{Visualization of the attention maps from our method. `Real' and `Prediction' indicates the ground-truth and predicted abnormal findings.}
    \label{fig:attention}
    % \vspace{-0.5\baselineskip}
\end{figure*}

\subsection{Performance comparison}

We conduct experiments on both the abnormal and complete set of the MIMIC-CXR dataset which consider the abnormal findings in reports and the complete reports, respectively.
As shown in \Cref{tbl:clinical_accuracy}, adding co-attention over medical concepts and dual decoders both improve the vanilla Hier-CNN-RNN model's clinical accuracy on the complete dataset. However, simply training the Hier-CNN-RNN model on the abnormal set would achieve better clinical accuracy. 
This shows the importance of addressing dataset bias. 
We also observe that the Hier-CNN-RNN model with a shuffled sentence order doesn't improve performance, which indicates the difficulty of addressing order bias during training of encoder-decoder models. 

Our CVSE model outperforms all baselines on clinical accuracy metrics, which demonstrates its capability to accurately report abnormal findings. 
Notably, CVSE achieves significant improvements on precision and recall. On the other hand, the baseline models will always miss abnormal findings thus leading to 0 precision and recall for many disease classes. More detailed results are included in the \textit{appendices}.

Refining the groups with mutual exclusivity rules further improves the performance of CVSE. We also report the automatic evaluation of NLG metrics. As shown in \Cref{tbl:clinical_accuracy}, CVSE achieves higher scores than other baselines on the abnormal set.\footnote{Models trained on the complete set can match the predominant normal findings thus leading to higher NLG metrics.} 

\subsection{Qualitative analysis}

We performed a human evaluation in which we sampled 20 images and asked a board-certified radiologist to give Likert scores (1 to 10) based on how closely the results generated by the model relate to the input images. The ground-truth obtained an average score of 7.85; our CVSE achieved a score of 6.35, higher than Hier-CNN-RNN trained on the abnormal set which obtained 6.15. The radiologist commented that Hier-CNN-RNN’s outputs were simpler predictions, with less details; meanwhile, CVSE covered more abnormalities but may included false information sometimes. 

In \Cref{fig:attention}, we visualize the attended regions on CXRs to investigate what part is important for reporting abnormal findings. We observe that our attention mechanism is able to detect relevant regions (e.g.~heart, left opacity, wires) to determine which abnormal findings reside in the CXRs.

\section{Conclusions}

In this paper, we study how to build assistive medical imaging systems that report abnormal findings on the medical images in the form of detailed descriptions. We formulate the problem as a cross-modal retrieval task and apply a metric learning-based method to align visual and semantic features (i.e.,~image regions and textual descriptions of abnormal findings) without explicit labels. Our experiments show that the retrieval-based method outperforms generation-based models by mitigating their weaknesses in generating repetitive sentences and bias toward normal findings. In the future, we will extend our method to other medical image datasets and explore transfer learning.

\section*{Acknowledgments}

This work was supported in part by the Office of the Assistant Secretary of Defense for Health Affairs through the Accelerating Innovation in Military Medicine Research Award program under Award No. W81XWH-20-1-0693, and NSF \#1750063.

% The acknowledgments should go immediately before the references. Do not number the acknowledgments section.
% Do not include this section when submitting your paper for review.

\bibliographystyle{acl_natbib}
% \bibliography{anthology,emnlp2020}
\bibliography{emnlp2020}

\appendix

\section{Implementation details}

\begin{table*}[t]
\centering
\small
\caption{Detailed Accuracy, precision and recall for different models.}
\label{tbl:detailed_accuracy}
\begin{tabularx}{\linewidth}{X r r r r r r}
\toprule
Model  & \multicolumn{3}{c}{CVSE + mutual exclusiveness} & \multicolumn{3}{c}{Hier-CNN-RNN (abnormal)} \\
\midrule
Disease  & Accuracy & Precision & Recall & Accuracy & Precision & Recall \\
\midrule
No Finding                 & \textbf{0.769}    & \textbf{0.346}     & \textbf{0.265}  & 0.766    & 0.336     & 0.259  \\
Enlarged Cardiomediastinum & 0.926    & \textbf{0.063}     & \textbf{0.060}  & \textbf{0.959}    & 0.000     & 0.000  \\
Cardiomegaly               & 0.801    & 0.512     & \textbf{0.606}  & \textbf{0.813}    & \textbf{0.570}     & 0.338  \\
Lung Lesion                & 0.921    & \textbf{0.192}     & \textbf{0.121}  & \textbf{0.943}    & 0.000     & 0.000  \\
Lung Opacity               & \textbf{0.692}    & \textbf{0.635}     & \textbf{0.237}  & 0.658    & 0.500     & 0.021  \\
Edema                      & 0.920    & 0.405     & \textbf{0.206}  & \textbf{0.927}    & \textbf{0.490}     & 0.084  \\
Consolidation              & 0.876    & \textbf{0.130}     & \textbf{0.181}  & \textbf{0.935}    & 0.079     & 0.006  \\
Pneumonia                  & \textbf{0.859}    & \textbf{0.364}     & \textbf{0.214}  & 0.855    & 0.306     & 0.154  \\
Atelectasis                & \textbf{0.773}    & \textbf{0.525}     & 0.320  & 0.599    & 0.284     & \textbf{0.469}  \\
Pneumothorax               & 0.964    & \textbf{0.073}     & \textbf{0.051}  & \textbf{0.977}    & 0.000     & 0.000  \\
Pleural Effusion           & \textbf{0.894}    & \textbf{0.640}     & 0.465  & 0.696    & 0.262     & \textbf{0.703}  \\
Pleural Other              & 0.962    & \textbf{0.145}     & \textbf{0.036}  & \textbf{0.968}    & 0.000     & 0.000  \\
Fracture                   & 0.917    & 0.063     & \textbf{0.050}  & \textbf{0.935}    & \textbf{0.072}     & 0.029  \\
Support Devices            & 0.808    & 0.348     & \textbf{0.321}  & \textbf{0.863}    & \textbf{0.752}     & 0.130  \\
Macro-Average              & \textbf{0.863}    & \textbf{0.317}     & \textbf{0.224}  & 0.850    & 0.261     & 0.157 \\
\bottomrule
\end{tabularx}
\vspace{-1em}
\end{table*}

\subsection{Mutual exclusive rules to refine groupings}

Though advanced sentence embedding methods allow for effective groupings of sentences in radiology reports describing similar clinical features, they fail to distinguish antonyms such as right vs. left because antonyms share highly similar contexts and are considered to be semantically similar by these embedding methods. For our purposes, however, it is important to distinguish some of the antonyms because they describe mutually exclusive image features. For example our grouping based on a sentence embedding results clustered these sentences in the same group:   
\textit{
\setlist{nolistsep}
\begin{itemize}[noitemsep]
    \item continued right lung volume loss.
    \item there is right lung volume loss again noted.
    \item right lung volume loss is again noted.
    \item there is volume loss of the left upper lung.
    \item left upper lobectomy changes including left lung volume loss.
    \item left upper lobe volume loss is present.
\end{itemize}
}
To separate those denoting right lung volume loss from those denoting left we wrote simple matching rules to identify selected words in sentences in the same group that are mutually exclusive and encode their occurrences as one-hot vectors. Then we applied the DBSCAN clustering method in the sklearn%
\footnote{\url{https://scikit-learn.org/stable/}} library to divide the group further into on average three subgroups based on the one-hot vector encoding. We considered six sets of mutually exclusive terms:
\setlist{nolistsep}
\begin{itemize}[noitemsep]
    \item right, left, bilateral.
    \item small, great$|$large.
    \item low, high.
    \item elevate$|$enlarge$|$increase$|$widen, shrink$|$decrease.
    \item improve$|$resolve$|$clear, worsen.
    \item mild, severe.
\end{itemize}

\subsection{Parameter settings}

We use PyTorch to implement all models and run them on 2 1080Ti GPUs. We resize all images into size of $512 \times 512$ for both models. For all experiments, we save the models that perform best on the validation set. For CVSE, we measure recall on validation set; for CNN-RNN models, we consider perplexity on validation set.

For CVSE we use an Adam optimizer with a learning rate 0.001 and training continues for 40 epochs. 
For all Hier-CNN-RNN models, we set the learning rate for encoder and decoder as $5e^{-6}$ and $2e^{-4}$, respectively. We train the models for 100 epochs. We use a VGG-19 model as the encoder, a 1-layer LSTM as the sentence decoder and a 2-layer LSTM as the word decoder. We observe slightly better performance from VGG-19 compared to DenseNet-121 for the generation models. For models that require medical concepts, we use SemRep (i.e.~a UMLS-based program released by NIH) to extract 93 highly frequent medical concepts from the training set.

\section{Experiments on MIMIC-CXR}
\subsection{Detailed clinically accuracy results on 14 diseases}
\Cref{tbl:detailed_accuracy} shows the detailed accuracy, precision and recall on all 14 diseases from our CVSE model with mutual exclusiveness rules and the Hier-CNN-RNN model trained on the abnormal set. Overall, CVSE outperforms Hier-CNN-RNN on the macro-average of accuracy, precision and recall. Notably, CVSE achieves higher recall on 12 out of 14 diseases with a comparative or higher precision. Meanwhile, Hier-CNN-RNN outputs 0 positive predictions on 4 disease types that are dominated by the negative findings, which shows its limited capability to generate diverse predictions.

% \section{Supplemental Material}
% \label{chapter6:sec:supplemental}

\end{document}